\documentclass{article} 

\usepackage{amsmath,amsfonts,bm}

\def\eqref#1{equation~\ref{#1}}

\def\1{\bm{1}}

\DeclareMathAlphabet{\mathsfit}{\encodingdefault}{\sfdefault}{m}{sl}
\SetMathAlphabet{\mathsfit}{bold}{\encodingdefault}{\sfdefault}{bx}{n}

\usepackage[dvipsnames]{xcolor}
\usepackage{hyperref}
\hypersetup{
    colorlinks=true,
	citecolor=MidnightBlue,
	linkcolor=OrangeRed,
	urlcolor=OrangeRed
}

\usepackage{iclr2027_conference,times}

\usepackage{hyperref}
\usepackage{url}
\usepackage{algorithm}
\usepackage{algorithmic}
\usepackage{amsmath}
\usepackage{amssymb}
\usepackage{subcaption}
\usepackage{multirow}
\usepackage{booktabs}
\usepackage{xcolor}     
\usepackage{colortbl}
\usepackage{newfloat}
\usepackage{amssymb}
\usepackage{pifont}
\usepackage{listings}
\usepackage{graphicx}
\usepackage{tabularx}
\newcolumntype{Y}{>{\centering\arraybackslash}X}
\usepackage[capitalise,nameinlink]{cleveref}
\crefname{figure}{Fig.}{Figs.}
\Crefname{figure}{Figure}{Figures}
\crefname{table}{Tab.}{Tabs.}
\Crefname{table}{Table}{Tables}

\title{SceneAE: Scene Generation and Reconstruction  in Queryable 3D-Anchored Latent Space}

\author{Shuofeng Sun\textsuperscript{1}, Wenzhao Zheng\textsuperscript{2}, Yunpeng Zhang\textsuperscript{3}, Yi Wu\textsuperscript{4}, Linqing Zhao\textsuperscript{1}, Haibin Yan\textsuperscript{1}\thanks{Corresponding author}\\
Beijing University of Posts and Telecommunications\textsuperscript{1}, University of California, Berkeley\textsuperscript{2}, \\ PhiGent Robotics \textsuperscript{3}, SynapX  \textsuperscript{4} \\
}
\iclrfinalcopy 
\begin{document}

\maketitle

\vspace{-8mm}
\begin{figure}[!h]
    \centering
    \includegraphics[width=1\linewidth]{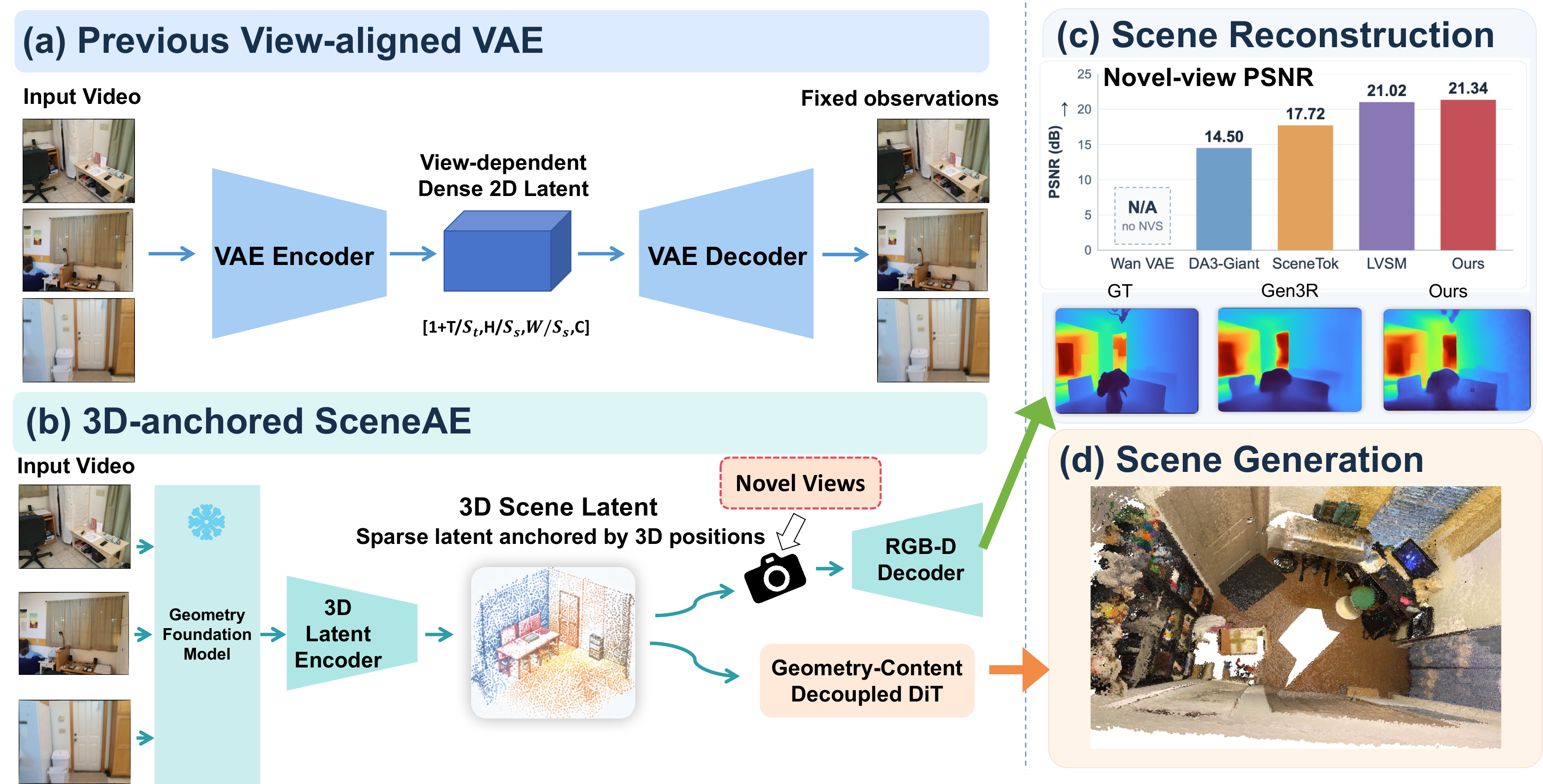}
     \vspace{-7mm}
    \caption{(a) Previous VAEs construct view-bound grid latent representations, which can only reconstruct fixed observations. (b) We construct a compact 1D latent representation anchored to explicit 3D positions, which can achieve efficient scene reconstruction and scene generation tasks.}
    \label{fig:teaser}
\end{figure}

\begin{abstract}
Recent video world models have increasingly adopted 3D memory mechanisms to integrate historical observations into a lightweight and persistent scene representation. This naturally raises a complementary question: if 3D representations can serve as an effective mechanism for memorizing observed scenes, can they also be directly used as the latent space for reconstructing and generating scenes? 
Motivated by this idea,  we represent a scene as a compact and unstructured 3D-anchored latent space (\textbf{SceneAE}).
Each latent combines an
  explicit 3D position with an implicit content feature and is spatially grounded to describe its local scene region, forming a
  sparse latent point cloud.
Unlike conventional video-VAE latents that describe how a scene appears from particular viewpoints, our
  representation directly describes the scene itself, making it naturally applicable to both reconstruction and generation. For
  reconstruction, once the scene has been generated, our persistent 3D representation can be projected and decoded into RGB-D observations by different target cameras, without the need to regenerate a video for each new camera trajectory. 
    For generation, we leverage the geometry-content separation to first generate camera-controlled 3D geometry and
  then synthesize content conditioned on it, reducing camera-appearance entanglement and improving camera
  controllability.
Experiments show that this 3D latent achieves superior
generation quality, computational cost, and camera controllability.


\end{abstract}

\section{Introduction}

  Recent video generation models~\citep{wan2025wan,hong2022cogvideo,yang2025cogvideox,sora}  encode videos into dense spatiotemporal latent grids and learn their
  distributions with diffusion transformers. Building on this paradigm, camera-controllable world models generate future
  observations from an initial view and a prescribed camera trajectory, often predicting RGB together with depth to improve spatial
  understanding. For example, 
  Gen3R~\citep{gen3r} jointly models the appearance latents of Wan VAE and the geometric features of VGGT~\citep{wang2025vggt}, while GLD~\citep{GLD}
  performs diffusion directly in the feature space of a geometry foundation model~\citep{da3}. These approaches combine scalable video-
  generation priors with explicit geometric information, producing compelling and geometry-aware RGB-D sequences.

Despite architectural differences, these methods use viewpoint-aligned latent grids that describe individual
  observations rather than the scene itself. Their outputs are therefore trajectory-specific videos that cannot be
  directly queried from novel viewpoints as shown in \Cref{fig:teaser}(a). Overlapping views also introduce latent redundancy, while appearance-
  centric representations make geometric consistency harder to learn.
SceneTok~\citep{scenetok} compresses the scene into compact unstructured 1D tokens and supports multi-view decoding, thereby alleviating the fixed-viewpoint limitation. However, its tokens lack explicit spatial localization, and the model must learn the scene region corresponding to each token through dense token-pixel interaction, which increases optimization difficulty as well as encoding and decoding costs. Our core idea is to anchor latent features at explicit 3D positions, guiding them to learn the corresponding local 3D regions and thereby forming a sparse latent point cloud, which is shown in \Cref{fig:teaser}(b). Unlike view-aligned dense grids, this representation describes the scene directly in a shared 3D coordinate system, so that the same physical region needs to be encoded only once and can be reused across different viewpoints, which avoids the redundancy of repeated encoding while effectively ensuring geometric consistency across viewpoints. Furthermore, due to the properties of the 3D representation, \textbf{for any target camera within the field of view}, the latents can be projected onto its image plane and decoded into per-pixel RGBD information through a deep learnable decoder, which allows our representation to directly perform the scene reconstruction task and thereby support fast world roaming.

Furthermore, existing camera-controllable generation methods typically inject camera features directly into view-aligned appearance latents. Since camera conditions and appearance features describe different factors, this implicit conditioning requires the model to infer geometric transformations, occlusion relationships, and the emergence of newly revealed regions while simultaneously generating texture, which easily leads to feature entanglement and makes camera control difficult. 3D memory methods such as GEN3C~\citep{ren2025gen3c}, Voyager~\citep{voyager}, and Mirage~\citep{mirage} have shown that more direct geometric control can be achieved when camera control acts directly on explicit scene geometry (e.g., a point cloud memory) rather than on appearance latents. However, the 3D memory in these methods is mainly used to store and query scene content that has already been observed.
Inspired by this, we further extend camera control over explicit geometry to future scene generation, and propose a two-stage Geometry-Content Decoupled DiT. Specifically, the future 3D anchor layout is first generated according to the target camera trajectory, and then the predicted point positions are used to provide spatial relationships through 3D RoPE to guide the generation of the corresponding latent content. This establishes a direct control path from camera trajectory to scene geometry and then to scene content, reducing camera-appearance entanglement and achieving more accurate trajectory following.

  Extensive experiments on multiple real-world datasets demonstrate that our method achieves superior scene reconstruction and generation quality with a substantially more
  compact latent space and lower computational cost. It consistently reconstructs geometrically coherent RGB-D observations across query viewpoints and more faithfully follows
  target camera trajectories.

\section{Related Work}
\textbf{Scene Latent Space.} 
Conventional image and video VAEs~\citep{wan2025wan,yang2025cogvideox} compress appearance into dense latent grids,
  but their limited geometric awareness can cause cross-view inconsistency. GLD~\citep{GLD} introduces stronger
  geometry by generating dense foundation-model features, at the cost of a high-dimensional latent space. Both
  remain viewpoint-aligned and redundantly encode overlapping regions. SceneTok~\citep{scenetok}
  uses compact, unstructured tokens for query-view decoding, but their lack of spatial grounding
  requires token-pixel interactions to learn token-region correspondence, increasing optimization difficulty
  and cost.

\textbf{Camera-Controllable Scene Generation.}
Recent video diffusion models~\citep{WVD,geometry_forcing} have been extended to camera-controllable scene generation by incorporating camera poses, ray embeddings, or geometric features into the
  generation process. Aether~\citep{aether} jointly models and generates image latents, depth latents, and camera pose latents. PRoPE~\citep{prope} encodes relative projective relationships directly in
  attention.  GLD~\citep{GLD} performs diffusion in the feature space of a geometry
  foundation model. Other methods maintain a persistent 3D memory: GEN3C~\citep{ren2025gen3c} renders an explicit point-cloud cache as conditioning,  and Mirage~\citep{mirage} stores and queries historical video latents in 3D space. 

\textbf{3D Scene Reconstruction.}  Neural scene representations, such as NeRF~\citep{nerf1,nerf2,nerf3} and 3D Gaussian Splatting~\citep{3dgs1,3dgs2,3dgs3}, encode scenes in 3D and support novel-view rendering, but typically require per-scene optimization and
  are not designed as latent spaces for generative modeling.  Therefore, some feed-forward reconstruction models~\citep{da3,worldmirror} efficiently and generally recover scene geometry from visual observations by predicting pixel-aligned 3D Gaussians. In addition, some methods~\citep{rayzer,scenetok,lvsm} reconstruct observations from arbitrary viewpoints by constructing implicit representations of the scene. 
  

\section{Method}

\subsection{3D Latent Representation}
We encode the video scene into a representation $R=\{P, Z\}=\{p_i, z_i\}_{i=1}^N$, where the explicit coordinates $p_i$ provide each latent with a stable spatial identity, and the implicit features $z_i$ describe the content of its corresponding local 3D region. Leveraging the properties of 3D point clouds, this representation reduces the redundancy of repeated observations across viewpoints, yielding a compact latent space while effectively supporting high-quality RGBD rendering from arbitrary viewpoints. 
\label{sec:method1}

\textbf{Encoder.}
Given $V$ posed RGB observations ($O={( I_v, K_v, T_v)}_{v=1}^{V}$, where $I_v\in R^{3\times H\times W}$,$K_v\in R^{4\times4}$ and $T_v\in R^{4\times4}$ represent the v-th RGB image, camera intrinsics, and the camera pose.), we aim to encode the observed scene into a sparse and view-independent 3D
  latent representation. We first apply a frozen Depth Anything 3-Base (DA3)~\citep{da3} model to extract multi-level feature maps ($F=[F_1,F_2....F_V]$) and predict depth maps ($D=[D_1,D_2...D_V]$). Each valid pixel
  is back-projected into the world coordinate system using $K$ and $T$ to obtain a global point cloud, and farthest-point sampling yields $N$ anchor positions $P=[p_1,p_2...p_N]$.  
  
\begin{figure}[t]
    \centering
    \includegraphics[width=0.9\linewidth]{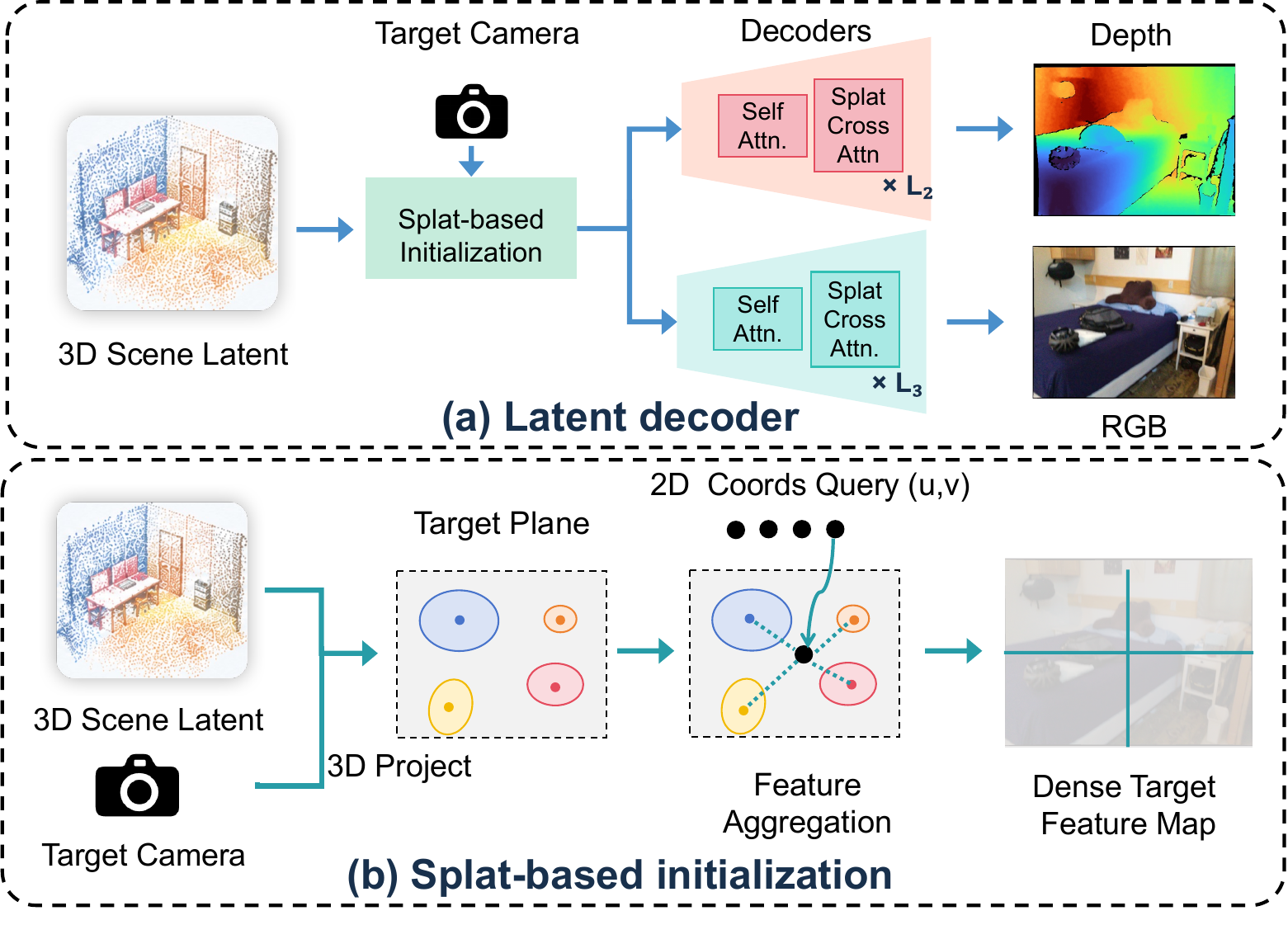}
    \vspace{-3mm}
    \caption{Overall Framework of the RGB-D Decoder. By leveraging the 3D properties of the latents, we can efficiently achieve RGBD rendering from arbitrary novel viewpoints.}
    \label{fig:vae}
    \vspace{-5mm}
\end{figure}
  For each latent, we project its 3D position into all visible input views and sample the corresponding DA3 features. These projected features are aggregated to initialize a latent query:
 \begin{equation}
 \label{eq:enc_query}
  q_i
  =
  \phi_x( p_i)
  +
  \phi_f\left(
  \operatorname{Agg}_{v\in \mathcal{V}_i}
  F_v\bigl(\pi_v( p_i)\bigr)
  \right),
  \end{equation}
  where $\phi_x (p_i)$ means the position embedding of the 3D coords. $\pi_v (p_i)$ projects the point $p_i$ into the v-th view using the corresponding camera parameters. Therefore, $F_v\bigl(\pi_v( p_i)\bigr)$ denotes the v-th view DA3 feature sampled at its projected location. Notably, instead of using all views, we select the top-$\mathcal{V}$ views for aggregation based on visibility and occlusion.

 In $E_{\theta}$, following BEVFormer~\citep{bevformer}, we construct deformable cross-attention with the DA3 feature maps based on the 3D coordinates of each latent, along with self-attention, for refinement.
  \begin{equation}
  \label{eq.enc}
  \{\mathbf z_i\}_{i=1}^{N}
  =
  E_{\theta}
  \left(
  \{q_i,p_i\}_{i=1}^{N},
  \{F_v\}_{v=1}^{V}
  \right).
  \end{equation}

  \textbf{Decoder.} 
  After constructing the non-grid latent representation, achieving RGBD reconstruction from arbitrary viewpoints remains a challenging problem. SceneTok decodes by having the pixels of the target view interact with all latents, which incurs substantial computational overhead. Given that our 3D latent representation is analogous to 3D Gaussians, for a target $v$-th view we first construct a corresponding downsampled grid $G_v\in R^{\frac{H}{S_s}\times \frac{W}{S_s}}$, where $S_s$ denotes the downsampling scale, and then splat each latent into the grid according to the projection relationship. For the j-th pixel in the grid, the top-$M_1$ nearest latents are selected based on 2D distance, and then the weight of each latent is computed from its distance, visibility, and latent features. 

  \begin{equation}
      s^v_{ij}=\psi(z_i, d^{2d}_v[ij]-\beta_{z}d^{z}_v(p_i)
  \end{equation}
where $s^v_{ij}$ mean the score of the $i$-th latent at the $j$-th pixel of the $v$-th view, $d^{2d}_v[ij]$ denotes the distance between the $j$-th pixel and the 2D coordinate onto which the $i$-th latent is splatted in the $v$-th view, and $d^{z}_v(p_i)$ denotes the depth of the $i$-th latent in the $v$-th view. This term is introduced so that latents with smaller depth receive higher weights, forming a soft z-buffer splatting.

  Then, these are aggregated into the initialization feature of the current $j$-th pixel in $v$-th view grid.
  \begin{equation}
  \label{eq:decoder_init}
      h^v_{j}=\sum_{n\in \mathcal{N}^v_{j}}s^v_{nj}\theta_{d}(z_n)
  \end{equation}
  where $h^v_{j}$ means the init feature, and $\mathcal{N}_j^v$ means the top-$M_1$ nearest latents set.

 This produces a low-resolution feature map $h$ for the target view. Unlike 3D Gaussian Splatting, we use splatting
  only for geometry-aware feature initialization rather than full-resolution rendering. This efficiently establishes 3D-to-2D
  correspondences at low resolution, while a deep decoder refines and upsamples the features to recover detailed RGB-D
  observations.

Finally, the RGB and depth predictions are obtained by the RGB Decoder $H_{\mathrm{rgb}}$ and Depth Decoder , in which these decoders consist of multiple layers of self-attention and cross-attention with the top-$M_2$ nearest latents under the view (splat cross attention). The overall framework of decoder is illustrated in \Cref{fig:vae}. More details are provided in \cref{sec:app_decoder}. 
\begin{equation}
\label{eq:decoder_rgb}
     \hat{\mathbf I}_v=H_{\mathrm{rgb}}(\{\mathbf h^v\},\{p_i,z_i\}_{i=1}^N),\hat{\mathbf D}_v=H_{\mathrm{depth}}(\{\mathbf h^v\},\{p_i,z_i\}_{i=1}^N)
\end{equation}

     

\textbf{Training.} The training losses are only the RGB and depth rendering losses.
\begin{equation}
    L_{latent}=\lambda_1L^{rgb}_{mse}+\lambda_2L^{rgb}_{grad}+\lambda_3L^{depth}_{mse}+\lambda_4L^{depth}_{grad}
\end{equation}

Furthermore, since the latent coordinates are also generated by the DiT at inference time, we apply small perturbations to the latent coordinates during the decoder training stage to enhance the robustness of the decoder to coordinate errors produced by the DiT.

\subsection{Geometry-Content Decoupled DiT}
\label{sec:method2}
\begin{figure}[t]
    \centering
    \includegraphics[width=1.0\linewidth]{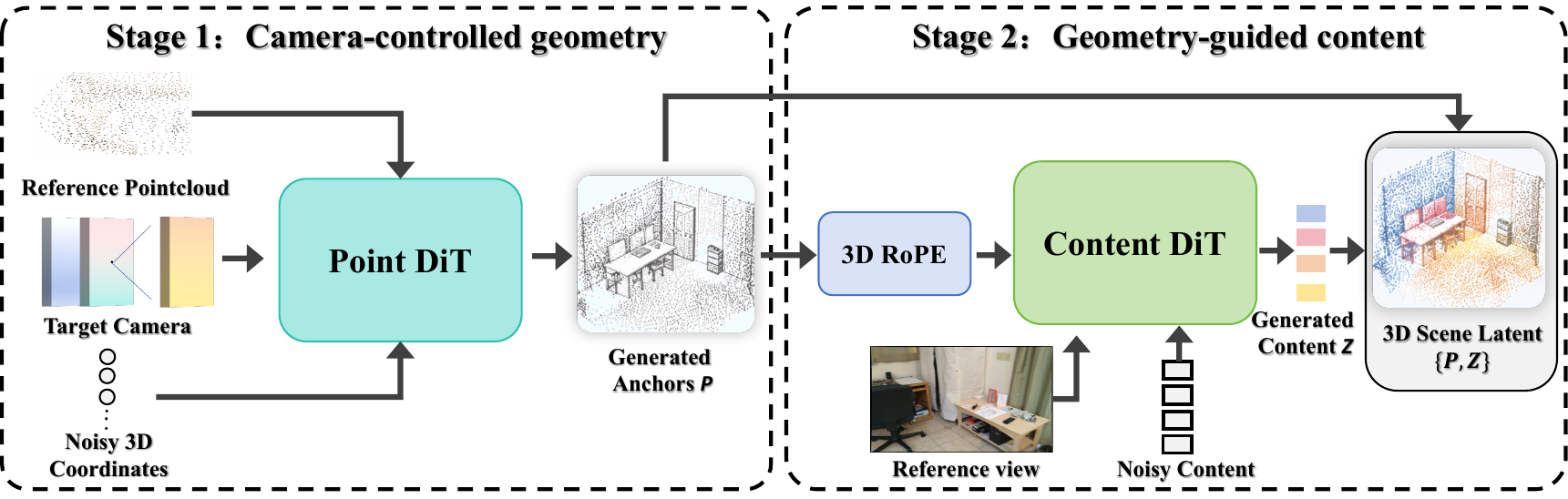}
    \vspace{-7mm}
    \caption{Overall Framework of the Geometry-Content decoupled DiT. We first control the generation of the future geometric layout via the camera, and then control the generation of texture content via the geometric layout, thereby achieving more accurate camera control.}
    \label{fig:dit}
    \vspace{-5mm}
\end{figure}

  Given the 3D latent state $R_{ref}=\{P_{ref}, Z_{ref} \}$ encoded from the reference frame and a target camera trajectory $T_{target}$, our goal is to generate the 3D positions ($P_{target}$) and
  content features ($Z_{target}$) of the target scene region. Existing methods commonly inject camera poses or rays into viewpoint-aligned video latents through feature addition, positional
  encoding, or attention. In these formulations, the camera trajectory remains an implicit conditioning signal, and the model must learn to translate it into changes in scene geometry,
  visibility, and spatial coverage. Because these geometric effects are expressed indirectly through appearance-oriented latents, accurate camera following remains challenging. In contrast, our
  3D latent representation provides an explicit geometric variable on which camera control can directly operate. We therefore first generate the spatial structure 
  according to the target trajectory and subsequently generate content features upon this geometry, the overall framework is shown in \Cref{fig:dit}. This establishes a structured control path
  from camera trajectory to scene geometry and then to scene content, enabling more accurate camera control.

  \textbf{Camera-Controlled Point Diffusion.}
We build a lightweight point diffusion model that generates future 3D positions from the reference point cloud and target camera poses.

  We first express all cameras relative to the reference camera and construct a sparse ray map for each target view. Each ray token
  contains its Plücker coordinates, image-plane location, trajectory index, and the depth information projected from the observed
  scene when available.

We then add noise to the input positions. To provide scene-aware information, we train using the flow path from the first-frame to the target point cloud. At timestep $t$, the noisy input is:
\begin{equation}
    \hat{P}^t=(1-t)\cdot P_{target}+t\cdot P_{ref} 
\end{equation}

The input noise is fed into a lightweight Point DiT for prediction. The Point DiT consists of self-attention on the input, cross-attention with the point cloud of the reference frame, and cross-attention with the ray token of the corresponding target camera after projection onto the target plane. As a result, the camera trajectory can directly constrain the spatial structure of the generated latents.

Finally, the model predicts the clean target layout $\hat{P}_{target}$ with an MSE loss.
\begin{equation}
    L_{point}=MSE(\theta_{point}(\hat{P}^t,P_{ref},T_{target}),P_{target})
\end{equation}

\textbf{Geometric-guided Content Diffusion.}
 The second-stage Content DiT generates content features on the geometry $\hat{P}_{target}$ produced by the first-stage Point DiT. We employ a Wan-style~\citep{wan2025wan} transformer consisting of adaptive LayerNorm self-attention, cross-attention to the reference latents $R_{ref}$, and
  feed-forward layers. In self- and cross-attention, we add 3D RoPE of the token position $\hat{P}_{target}$ for geometric association and camera control. Training follows:
  \begin{equation}
  \label{eq:content_noise}
      \hat{Z}^t=(1-t)\cdot Z_{target}+t\cdot \sigma
  \end{equation}
  \begin{equation}
  \label{eq:content_loss}
      L_{content}=MSE(\theta_{content}(\hat{Z}^t,R_{ref}, \hat{P}_{target}),Z_{target}-\hat{Z}^t)
  \end{equation}

\begin{table}[t]
\centering
\caption{Quantitative Comparison of Appearance Generation. We report the PSNR, SSIM, Subject Consistency, Background Consistency, Multi-view Consistency from VBench. We also report the GFLOPs and ms per step to evaluate the efficiency of the models.
}
    \vspace{-2mm}
\label{tab1}
\resizebox{1.0\linewidth}{!}{
 \setlength{\tabcolsep}{2.5pt}
\begin{tabular}{c|c|cccccccccc|cc}
\toprule
\multirow{2}{*}{\rotatebox[origin=c]{90}{\scriptsize \textbf{Cond.}}} & \multirow{2}{*}{Method} &
     
    \multicolumn{5}{c}{DL3DV-10K}&\multicolumn{5}{c|}{RealEstate10K}& ~~~\multirow{2}{*}{GFLOPs}~~~& \multirow{2}{*}{ms} \\
    \cmidrule(lr){3-7}\cmidrule(lr){8-12}
    &&~~~PSNR$\uparrow$&SSIM$\uparrow$&Sub$\uparrow$&Bg$\uparrow$&MVC$\uparrow$&PSNR$\uparrow$&SSIM$\uparrow$&Sub$\uparrow$&Bg$\uparrow$&MVC$\uparrow$\\
    \midrule
     \multirow{5}{*}{\rotatebox[origin=c]{90}{\textbf{1-view}}} &Aether&14.1&0.36&0.81&0.84&0.19&15.4&0.47&0.82&0.89&0.21&48874&364\\
     &SceneTok&-&-&-&-&-&15.3&0.48&0.84&0.91&0.24&-&-\\
     &Gen3R&14.7&0.39&0.85&0.88&0.21&15.8&0.53&0.86&0.92&0.24&18105&174\\
     &GLD&15.6&0.45&0.87&0.91&0.22&16.9&0.58&0.89&0.94&0.28&23203&285\\
     \rowcolor{blue!10}
     &Ours&15.4&0.45&0.88&0.93&0.28&17.1&0.60&0.90&0.94&0.33&8348&134\\
     \midrule
     \multirow{5}{*}{\rotatebox[origin=c]{90}{\textbf{2-view}}} &Aether&15.7&0.42&0.82&0.84&0.20&16.8&0.55&0.82&0.88&0.23&48874&364\\
     &SceneTok&-&-&-&-&-&17.1&0.57&0.84&0.90&0.25&-&-\\
&Gen3R&16.3&0.46&0.86&0.89&0.24&17.5&0.59&0.85&0.92&0.26&18105&174\\
&GLD&18.1&0.53&0.87&0.92&0.23&19.4&0.67&0.88&0.93&0.30&23203&285\\
\rowcolor{blue!10}
     &Ours&17.9&0.52&0.88&0.92&0.28&19.8&0.68&0.92&0.95&0.34&9975&145\\
    
\bottomrule
\end{tabular}
}
    \vspace{-6mm}
\end{table}

  \textbf{Two-stage Training.} Since camera-controllable content generation relies on 3D RoPE, relatively accurate 3D positions are required. Therefore, we adopt a two-stage training paradigm: the Point DiT is trained first, and only then is the Content DiT trained, rather than training them jointly. Details of the Point DiT and Content DiT are in \cref{sec:app_point,sec:app_content}.

  \section{Experiments}
  \subsection{Training Details}
  \textbf{Datasets.} 
We train SceneLatent from scratch on DL3DV~\citep{dl3dv}, RealEstate10K~\citep{re10k}(RE10K), and ScanNet~\citep{scannet}. For a fair comparison, we construct each sample from nine views (matching the 4N+1 frame requirement of Wan, Gen3R, etc., while approaching the 8-frame design used by GLD), and then train the scene generation task under both the 1-view setting with a given first frame and the 2-view setting with given first and last frames. 

\begin{table}[t]
\centering
\caption{Quantitative Comparison of Geometry Generation. We report the average accuracy, completeness, and Chamfer Distance. The evaluation GT for Gen3R is provided by VGGT~\citep{wang2025vggt}, while the evaluation GT for ours and GLD is provided by DA3~\citep{da3}. This avoids the influence of performance differences among geometry foundation models.
}
    \vspace{-2mm}
\label{tab2}
\resizebox{0.9\linewidth}{!}{
 \setlength{\tabcolsep}{2.5pt}
\begin{tabular}{c|c|cccccc}
\toprule
\multirow{2}{*}{\rotatebox[origin=c]{90}{\scriptsize \textbf{Cond.}}} & \multirow{2}{*}{Method} &
     
    \multicolumn{3}{c}{DL3DV-10K}&\multicolumn{3}{c}{RealEstate10K}\\
    \cmidrule(lr){3-5} \cmidrule(lr){6-8}
&&~~~Accuracy$\downarrow$~~~&~~~Completeness$\downarrow$~~~&~~~CD$\downarrow$~~~&~~~Accuracy$\downarrow$~~~&~~~Completeness$\downarrow$~~~&~~~CD$\downarrow$~~~\\
\midrule

     \multirow{3}{*}{\rotatebox[origin=c]{90}{\textbf{1view}}} &Gen3R&1.74&1.92&2.01&0.75&0.91&0.96\\
     &GLD&0.87&1.18&1.04&0.36&0.49&0.46\\
    
     &Ours&0.34&0.55&0.44&0.13&0.22&0.18\\
     
     \midrule
     \multirow{3}{*}{\rotatebox[origin=c]{90}{\textbf{2view}}} 
      &Gen3R&1.23&1.46&1.38&0.59&0.75&0.74\\
     &GLD&0.63&0.91&0.85&0.27&0.44&0.39\\
   
     &Ours&0.21&0.46&0.30&0.07&0.10&0.08\\
\bottomrule
\end{tabular}
}
    \vspace{-4mm}
\end{table}

\textbf{Training Details.}
We use DepthAnything V3-Base as the backbone to extract feature maps and depth for the input videos.
The image resolution is set to 518×376, and the number $N$ of 3D latents is set to 3000. The training optimizer hyperparameters follow GLD, using AdamW with a learning rate of 5e-5 and a batch size of 40. SceneLatent is trained for 100k iterations on 8 A800 GPUs. Although our method uses less training time and fewer training datasets than methods such as GLD and Gen3R, it still achieves competitive performance.

\begin{table}[t]
  \centering
  \begin{minipage}[b]{0.485\textwidth}
    \centering
    \caption{Trajectory following. AUC@30 and ATE (Absolute Trajectory Error) are computed from poses estimated by VGGT.}
    \label{tab3}
  \end{minipage}\hfill
  \begin{minipage}[b]{0.485\textwidth}
    \centering
    \caption{Memory ability. We reverse each trajectory to form a round trip, and compare corresponding forward and return views.}
    \label{tab4}
  \end{minipage}
  \par\vspace{-1mm}
  \footnotesize
  \begin{minipage}[t]{0.485\textwidth}
    \centering
    \setlength{\tabcolsep}{3pt}
    \begin{tabularx}{\linewidth}{c|YYYY}
      \toprule
      \multirow{2}{*}{Method} & \multicolumn{2}{c}{DL3DV} & \multicolumn{2}{c}{RE10K} \\
      & AUC@30$\uparrow$ & ATE$\downarrow$ & AUC@30$\uparrow$ & ATE$\downarrow$ \\
      \midrule
      Aether & 0.48 & 0.23 & 0.51 & 0.22 \\
      Gen3R & 0.59 & 0.19 & 0.57 & 0.20 \\
      GLD & 0.64 & 0.15 & 0.67 & 0.15 \\
      \midrule
      \rowcolor{blue!10}
      Ours & \textbf{0.76} & \textbf{0.10} & \textbf{0.72} & \textbf{0.09} \\
      \bottomrule
    \end{tabularx}
  \end{minipage}\hfill
  \begin{minipage}[t]{0.485\textwidth}
    \centering
    \setlength{\tabcolsep}{3pt}
    \begin{tabularx}{\linewidth}{c|YYY}
      \toprule
      \multirow{2}{*}{Method} & \multicolumn{2}{c}{RGB} & Depth \\
      & PSNR$\uparrow$ & SSIM$\uparrow$ & Accuracy$\downarrow$ \\
      \midrule
      Aether & 17.1 & 0.58 & 0.18 \\
      Gen3R & 17.3 & 0.60 & 0.12 \\
      GEN3C & 29.8 & 0.90 & 0.06 \\
      \midrule
      \rowcolor{blue!10}
      Ours & \textbf{30.7} & \textbf{0.93} & \textbf{0.04} \\
      \bottomrule
    \end{tabularx}
  \end{minipage}
  \vspace{-7mm}
\end{table}

\subsection{3D Scene Generation}
\textbf {Appearance Generation.}  We evaluate the RGB generation performance on DL3DV Eval and RealEstate10K test, reporting mainly PSNR, SSIM~\citep{ssim}, as well as Subject Consistency, Background Consistency, Multi-View Consistency from VBench~\citep{vbench}, with comparison methods including Aether~\citep{aether}, SceneTok~\citep{scenetok}, Gen3R~\citep{gen3r} and GLD~\citep{GLD}. In addition, we report the GFLOPs and time (ms) per inference step for all evaluated methods. The results are shown in \Cref{tab1}. As can be seen, owing to the adaptability of our splat-based decoder to the 3D representation and the precise camera control brought by the geometry-content decoupled DiT, our method achieves the best generation quality even with an extremely sparse 3D latent representation. More importantly, benefiting from the sparsity of the latent representation, our computational overhead in GFLOPs and inference time is only half that of the state-of-the-art method GLD. This fully demonstrates that 3D representations can also be effectively applied to future scene generation tasks.

\textbf{Geometry Generation.}  
We evaluate depth generation on DL3DV Eval and the RealEstate10K test set using accuracy, completeness, and Chamfer
  distance. To avoid conflating generation quality with differences among geometry foundation models, each method is
  evaluated against the depth produced by its own geometry model: VGGT for Gen3R and DA3 for ours and GLD. As shown
  in \Cref{tab2}, our method substantially outperforms existing approaches, benefiting from latents anchored to
  explicit 3D coordinates and the resulting geometric prior throughout generation.

\textbf{Trajectory Following.} In this section, we test the trajectory following capability of the model during generation. Specifically, for the predicted videos, we compute the poses using VGGT and compare them with the ground-truth poses. We report AUC@30  and ATE (Absolute Trajectory Error), and the results are shown in \Cref{tab3}. As can be seen, thanks to the geometry-content decoupled DiT design, we achieve more precise camera control, yielding stronger trajectory following.

\begin{figure}[t]
    \centering
    \includegraphics[width=1.0\linewidth]{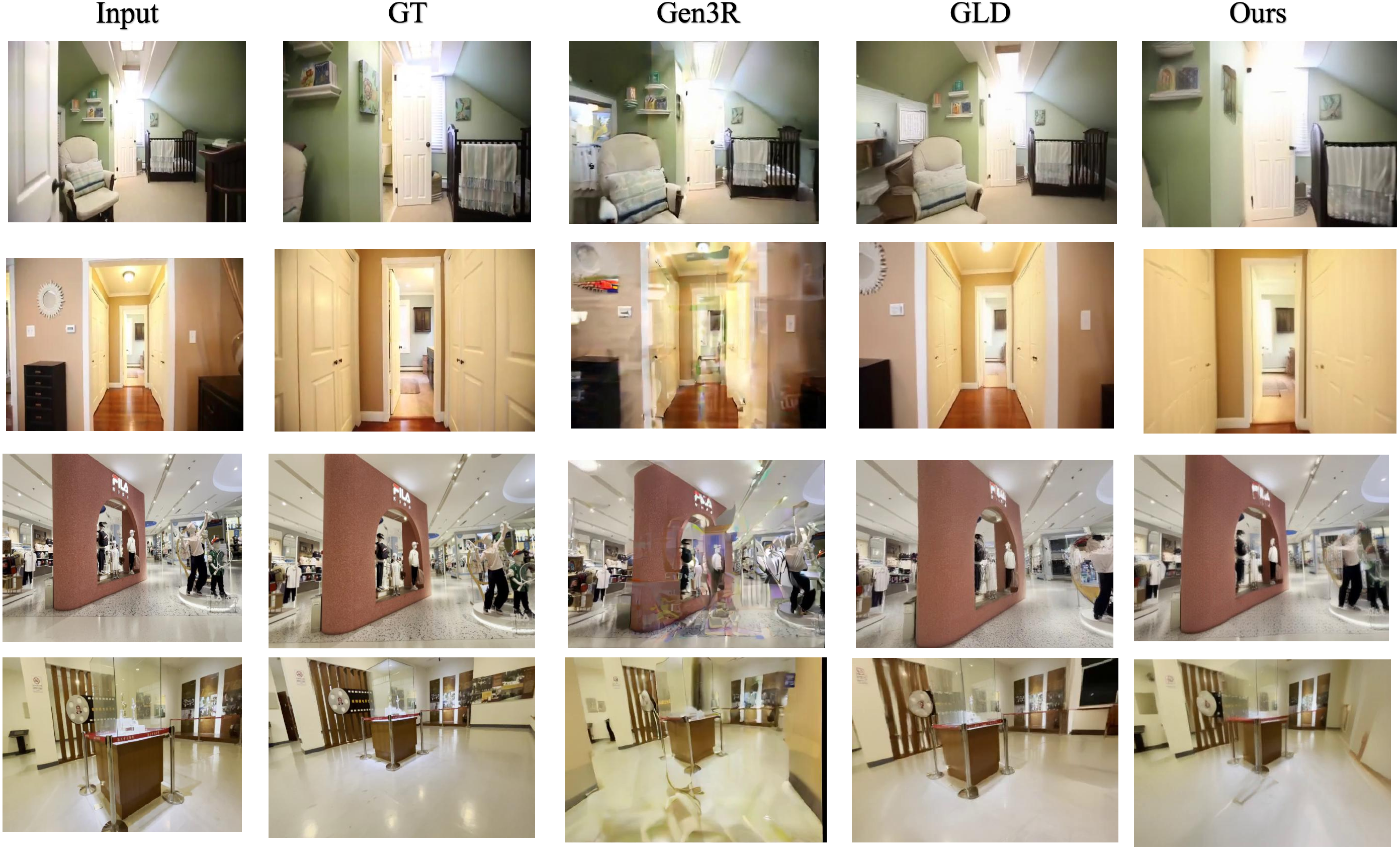}
    \vspace{-7mm}
    \caption{Qualitative Comparison of Different Methods under the 1-view setting.}
    \label{fig:vis_gen}
    \vspace{-6mm}
\end{figure}
\textbf{Memory Ability.} Memory is essential for a world model to remain consistent when revisiting previously observed regions. We evaluate
  this ability by concatenating each test trajectory with its reversed copy to form a round-trip sequence. We compare
  corresponding forward and return RGB views using PSNR and SSIM, and depth views using Accuracy. As shown in
  \Cref{tab4}, GEN3C outperforms Gen3R by using an explicit 3D point-cloud memory. Our latent memory also forms
  an implicit point cloud: nearby latent coordinates tend to carry consistent content, enabling strong round-trip
  consistency comparable to explicit 3D memory.

\textbf{Visualizations of RGB Generation.} In \Cref{fig:vis_gen}, we visualize the last generated frame under the 1-view setting for different methods. More visualizations are provided in \cref{sec:app_gen}.
\begin{table}[t]
  \centering
  \caption{Quantitative comparison of RGB reconstruction. We report PSNR and SSIM for reconstructing the input views (Input Reconstruction) and for rendering 20 other viewpoints of the scene from 9 context frames (Novel View Synthesis). ``--'' denotes not supported or not reported.}
  \vspace{-2mm}
  \label{tab5}
  \footnotesize
  \setlength{\tabcolsep}{6pt}
  \begin{tabular}{c|cccc|cccc}
    \toprule
    \multirow{3}{*}{Method} & \multicolumn{4}{c|}{Input Reconstruction} & \multicolumn{4}{c}{Novel View Synthesis} \\
    \cmidrule(lr){2-5}\cmidrule(lr){6-9}
    & \multicolumn{2}{c}{DL3DV} & \multicolumn{2}{c|}{RE10K} & \multicolumn{2}{c}{DL3DV} & \multicolumn{2}{c}{RE10K} \\
    & PSNR$\uparrow$ & SSIM$\uparrow$ & PSNR$\uparrow$ & SSIM$\uparrow$ & PSNR$\uparrow$ & SSIM$\uparrow$ & PSNR$\uparrow$ & SSIM$\uparrow$ \\
    \midrule
    Wan-VAE & 26.1 & 0.79 & 30.9 & 0.88 & -- & -- & -- & -- \\
    GLD & 26.7 & 0.84 & 31.1 & 0.91 & -- & -- & -- & -- \\
    DA3-Giant & -- & -- & -- & -- & 13.4 & 0.35 & 14.4 & 0.52 \\
    LVSM & -- & -- & -- & -- & 17.8 & 0.53 & 21.0 & 0.71 \\
    SceneTok & 19.3 & 0.61 & 24.1 & 0.81 & 16.1 & 0.46 & 17.7 & 0.59 \\
    \midrule
    \rowcolor{blue!10}
    Ours & 22.2 & 0.68 & 26.3 & 0.84 & \textbf{18.3} & \textbf{0.55} & \textbf{21.3} & \textbf{0.73} \\
    \bottomrule
  \end{tabular}
  \vspace{-7mm}
\end{table}

\begin{figure}[t]
    \centering
    \includegraphics[width=0.95\linewidth]{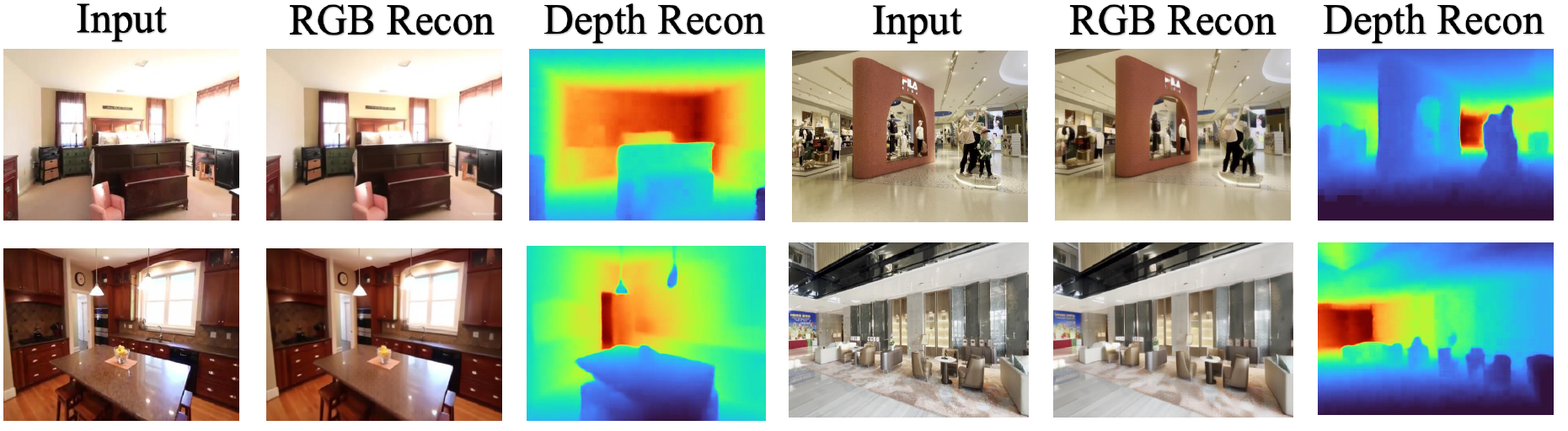}
    \vspace{-3mm}
    \caption{Visualization of the Latent Encoder-Decoder Reconstruction.}
    \label{fig:vis_recon}
    \vspace{-5mm}
\end{figure}

\begin{table}[t]
  \centering
  \begin{minipage}[b]{0.485\textwidth}
    \centering
    \caption{Depth reconstruction on DL3DV.}
    \vspace{-3mm}
    \label{tab:depth_recon}
  \end{minipage}\hfill
  \begin{minipage}[b]{0.485\textwidth}
    \centering
    \caption{Cross-view depth consistency.}
    \vspace{-3mm}
    \label{tab6}
  \end{minipage}
  \par\vspace{-1mm}
  \footnotesize
  \begin{minipage}[t]{0.485\textwidth}
    \centering
    \setlength{\tabcolsep}{3pt}
    \begin{tabularx}{\linewidth}{c|YYY}
      \toprule
      Method & Accuracy$\downarrow$ & Completeness$\downarrow$ & CD$\downarrow$ \\
      \midrule
      Aether & 0.74 & 0.96 & 0.79 \\
      Gen3R & 0.51 & 0.81 & 0.63 \\
      \midrule
      \rowcolor{blue!10}
      Ours & \textbf{0.12} & \textbf{0.35} & \textbf{0.20} \\
      \bottomrule
    \end{tabularx}
  \end{minipage}\hfill
  \begin{minipage}[t]{0.485\textwidth}
    \centering
    \setlength{\tabcolsep}{3pt}
    \begin{tabularx}{\linewidth}{c|YYYY}
      \toprule
      Method & Median$\downarrow$ & P90$\downarrow$ & 1\%$\uparrow$ & 10\%$\uparrow$ \\
      \midrule
      DA3-Base & 8.55\% & 47.55\% & 9.00\% & 54.58\% \\
      Gen3R & 8.65\% & 47.95\% & 8.92\% & 53.96\% \\
      \midrule
      \rowcolor{blue!10}
      Ours & \textbf{8.05\%} & \textbf{46.15\%} & \textbf{10.25\%} & \textbf{55.92\%} \\
      \bottomrule
    \end{tabularx}
  \end{minipage}
  \vspace{-8mm}
\end{table}

\subsection{3D SCENE RECONSTRUCTION}

\textbf{Novel View Synthesis.} A key property of our representation is that the encoded scene can be queried from viewpoints that are not observed in the input. Using 9 frames as context, we render 20 other viewpoints of the scene and report PSNR and SSIM in the right part of \Cref{tab5}. Dense view-aligned latents such as Wan-VAE and GLD can only decode the encoded observations and therefore do not support this setting. Our method outperforms the feed-forward Gaussian model DA3-Giant as well as the query-based SceneTok and LVSM on both DL3DV and RE10K. Since each latent is anchored to an explicit 3D position, projecting the latents onto a novel camera directly provides geometrically consistent correspondences for decoding, rather than requiring the decoder to infer them implicitly.

\textbf{Input Reconstruction.}  The left part of \Cref{tab5} compares the RGB reconstruction performance of different latent encoders and decoders. Wan-VAE and GLD adopt dense, view-aligned grids, which retain more visual information and higher fidelity. Like ours, SceneTok uses unstructured 1D tokens to support novel view reconstruction, but its latents must learn their responsible regions from scratch and require full interaction with pixels, resulting in higher optimization difficulty and greater computational overhead. By explicitly anchoring latents to 3D positions and restricting interaction to spatially relevant local regions, our method achieves higher reconstruction performance while significantly reducing computational overhead. Compared with SceneTok at the same resolution, GFLOPs drop from 27,096 to 13,171, and batch latency decreases from 4,020 ms to 1,623 ms. \Cref{fig:vis_recon} visualizes the RGB-D reconstructions of the input images.

\begin{figure}[t]
    \centering
    \includegraphics[width=0.86\linewidth]{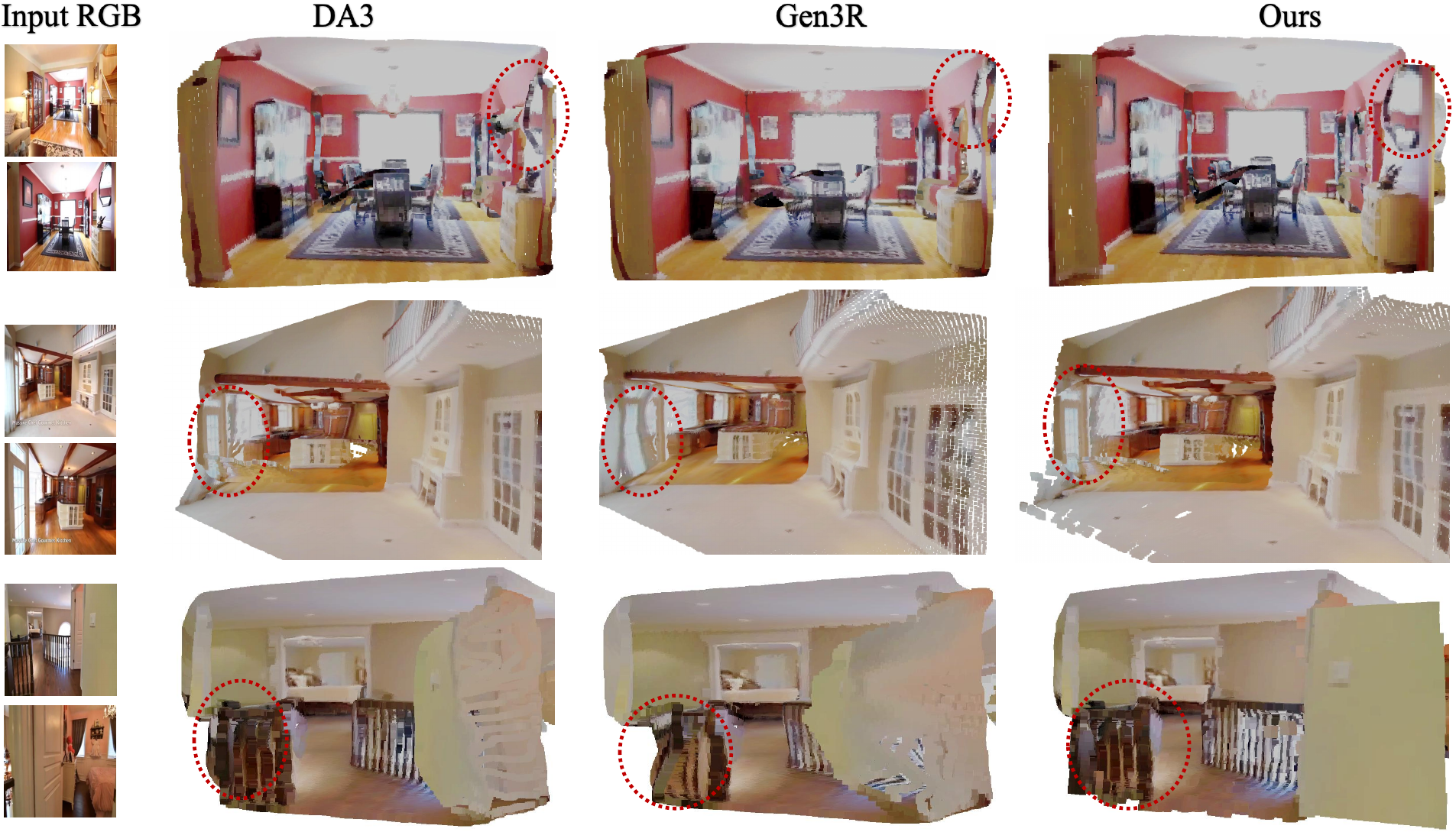}
    \vspace{-3.5mm}
    \caption{Qualitative comparison of geometry reconstruction. Red circles mark distorted regions.}
    \label{fig:geo_recon}
    \vspace{-7.5mm}
\end{figure}

\textbf{Geometry Reconstruction.} We evaluate depth reconstruction on DL3DV in \Cref{tab:depth_recon}. Our method reduces the Chamfer Distance from 0.63 (Gen3R) and 0.79 (Aether) to 0.20, since the explicit 3D anchors provide a strong geometric prior for depth decoding. We further measure the depth projection consistency between adjacent views in \Cref{tab6}, where our method also surpasses Gen3R and DA3-Base, indicating more consistent scene geometry across views. As shown in \Cref{fig:geo_recon}, our method faithfully recovers object structures, whereas DA3 and Gen3R may bend or distort objects.

\begin{table}[!t]
  \centering
  \begin{minipage}[b]{0.485\textwidth}
    \centering
    \caption{Ablation of latent autoencoder.}
    \vspace{-3mm}
    \label{tab7}
  \end{minipage}\hfill
  \begin{minipage}[b]{0.485\textwidth}
    \centering
    \caption{Ablation of DiT designs.}
    \vspace{-3mm}
    \label{tab8}
  \end{minipage}
  \par\vspace{-1mm}
  \footnotesize
  \begin{minipage}[t]{0.485\textwidth}
    \centering
    \setlength{\tabcolsep}{3pt}
    \renewcommand{\arraystretch}{1.28}
    \begin{tabularx}{\linewidth}{YYY|Y}
      \toprule
      Jitter & Low-Res & Splat-CA & PSNR$\uparrow$ \\
      \midrule
      \ding{55} & \ding{55} & \ding{55} & 24.1 \\
      \ding{55} & \ding{55} & \ding{51} & 24.7 \\
      \ding{55} & \ding{51} & \ding{55} & 25.9 \\
      \ding{55} & \ding{51} & \ding{51} & 26.3 \\
      \midrule
      \rowcolor{blue!10}
      \ding{51} & \ding{51} & \ding{51} & \textbf{26.5} \\
      \bottomrule
    \end{tabularx}
  \end{minipage}\hfill
  \begin{minipage}[t]{0.485\textwidth}
    \centering
    \setlength{\tabcolsep}{4pt}
    \renewcommand{\arraystretch}{1.1}
    \begin{tabularx}{\linewidth}{l|X|c}
      \toprule
      Setting & Variant & PSNR$\uparrow$ \\
      \midrule
      \multirow{3}{*}{(a) Generation} & Joint & 15.5 \\
      & Content-Only & 16.0 \\
      & \cellcolor{blue!10}Decoupled & \cellcolor{blue!10}\textbf{17.1} \\
      \midrule
      \multirow{3}{*}{(b) Camera Inj.} & Content-Only & 16.2 \\
      & Both & 16.4 \\
      & \cellcolor{blue!10}Geometry-Only & \cellcolor{blue!10}\textbf{17.1} \\
      \bottomrule
    \end{tabularx}
  \end{minipage}
  \vspace{-6mm}
\end{table}

\subsection{Ablation Study}
\textbf{Components of the Latent Encoder-Decoder.} \Cref{tab7} ablates three components of the latent encoder-decoder on RE10K. Here, \textbf{Jitter} denotes perturbing the latent coordinates during the decoder stage. \textbf{Low-Res} denotes the decoding paradigm that constructs a low-resolution dense feature map via splat-based initialization and then performs feature refinement through a subsequent deep decoder.  Rendering at the original resolution as in 3D Gaussian splatting performs worse, because our latents are much sparser than pixel-aligned Gaussians.
\textbf{Splat-CA} denotes the splat cross-attention in the decoder: unlike SceneTok, where each pixel interacts with all latents, our design lets each pixel interact only with the nearest few latents splatted onto the plane.

\textbf{Different DiT Design.} \Cref{tab8} evaluates alternative DiT designs on RE10K. In (a), Joint generates positions and content jointly,
  Content-Only generates content and regresses positions afterward, and Decoupled first generates geometry and then
  conditions content generation on it. The decoupled design performs best, showing that explicit geometric priors
  facilitate content synthesis. In (b), injecting the camera condition only into geometry generation outperforms
  injecting it into both geometry and content, confirming that camera-appearance entanglement harms generation
  quality.

\section{Conclusion}
We presented SceneAE, a sparse and scene-centric 3D latent representation for both scene reconstruction and generation. Unlike view-aligned dense grid latents, each latent couples an explicit 3D anchor position with an implicit content feature, so that every physical region is encoded only once and shared across viewpoints. A splat-based decoder renders this compact representation into RGB-D observations from arbitrary cameras without regenerating a video for each trajectory. Our Geometry-Content Decoupled DiT further generates the 3D anchor layout from the camera trajectory before synthesizing content, establishing an explicit control path from camera to geometry to content. Experiments show that SceneAE achieves strong generation quality, superior geometric consistency and camera controllability, at about half the computational cost of dense latents. We hope this work encourages the use of 3D representations as the latent space of video world models.

\bibliography{iclr2027_conference}
\bibliographystyle{iclr2027_conference}

\newpage
\appendix

{\Large\bfseries Appendix}

\section{Implementation Details}
\label{sec:app_impl}
\subsection{Latent Encoder}
\label{sec:app_encoder}
 We provide further details of the latent encoder ($E_\theta$) introduced in \cref{eq.enc} in the main paper. Given $N$ sparse latent
  positions ($P=\{p_i\}_{i=1}^{N}$) and $V$ posed RGB observations, the encoder associates each latent
  with a 64-dimensional content feature $z_i$. The latent coordinates remain fixed throughout encoding: the
  encoder only gathers and transforms visual evidence at these explicit 3D locations. This preserves a clear
  separation between the geometric support $P$ and the learned content $Z$.

  \textbf{Query Initialization.} We use a frozen DA3-Base backbone and extract 768-dimensional feature maps from all (L=12) transformer layers. For
  each latent $p_i$, we project it into every input view and retain projections that lie in front of the
  camera and inside the image boundary. To reduce feature contamination caused by occlusion or inaccurate
  correspondence, we compare the projected anchor depth $d_{iv}^{proj}$ with the DA3 depth sampled at the
  same image location.
  
  We select the view with the smallest depth discrepancy and sample its features at the anchor
  projection. The features are layer-averaged and added to a Fourier coordinate encoding (\cref{eq:enc_query}).

\textbf{Dual-path Deformable Cross Attention.}
Since we need to decode RGBD observations from the 3D latent representation, this involves a trade-off in aggregating multi-view features. Specifically, multi-view aggregation requires balancing feature reliability and scene coverage. Using only the views with the highest depth consistency yields accurately aligned local texture appearance, but may lack complementary geometry information visible from other views. Conversely, aggregating a large number of views can expand spatial coverage and enhance geometric understanding, but may also introduce conflicting texture features from occluded regions or incorrect correspondences. We therefore design two complementary deformable cross-attention paths. The Main path reads only the views with the highest depth consistency, prioritizing reliable local texture evidence, while the Multi-View path introduces more visible views layer by layer, supplementing geometric context.
\begin{equation}
    q_i^{l,main}=DA(q_i^l,\hat{F}^i,\hat{F}^i)
\end{equation}
\begin{equation}
    q_i^{l,mv}=DA(q_i^l,[F_1,F_2...F_V],[F_1,F_2...F_V])
\end{equation}
where $q_i^l$ denotes the $i$-thlatent feature input at layer $l$, $\hat{F}^i$ denotes the view with the highest depth consistency with the $i$-th latent, and DA denotes deformable cross-attention, which samples four points around the projected location in the input view for interaction.

 The two outputs are then fused to dynamically balance texture and geometric information.
 \begin{equation}
  \bar{\mathbf q}_i^{\,l}
  =
  q_i^{l}
  +
  \phi_{\mathrm{fuse}}
  \left(
  \left[
  q_i^{l,\mathrm{main}},
  q_i^{l,\mathrm{mv}}
  \right]
  \right).
  \end{equation}

\textbf{Self-Attention with 3D Rope}
After the dual-path deformable cross-attention, we employ global self-attention with 3D RoPE to enlarge the receptive field of the latents.
\begin{equation}
  \widetilde{\mathbf q}^{\,l}
  =
  \bar{\mathbf q}^{\,l}
  +
  \operatorname{SelfAttn}_{\mathrm{3D\text{-}RoPE}}
  \left(
  \operatorname{Norm}(\bar{\mathbf q}^{\,l}),
  \mathbf P
  \right),
  \end{equation}
  Therefore, the explicit coordinates serve two complementary roles in the encoder: on the one hand, they determine from which locations in the image the cross-attention should sample features, and on the other hand, they provide 3D spatial relationships for the latent self-attention

\subsection{Latent Decoder}
\label{sec:app_decoder}
\textbf{Splat Cross Attention}
Although the feature map initialized by \cref{eq:decoder_init} may be correctly assigned texture and geometric information, a single aggregation may be insufficient for occluded and sparsely covered regions. We retain the (K=64) nearest 3D latents for each pixel and perform local cross-attention in subsequent blocks. Each image token interacts only with its local latents.

\textbf{RGB Reconstruction}
Here we provide the specific details of \cref{eq:decoder_rgb}. We first refine the initialized feature map h (\cref{eq:decoder_init}) through self-attention and splat cross-attention, and then restore it to the image resolution via a two-stage PixelShuffle head. The first stage upsamples the patch grid by a factor of 7, and the second stage upsamples it by a factor of 2, ultimately producing the RGB image.

\textbf{Depth Reconstruction}
Here we provide the specific details of \cref{eq:decoder_rgb}. Unlike RGB reconstruction, since the 3D latent representation has explicit 3D coordinates, it can be leveraged to provide a depth prior. For each query viewpoint, we estimate a robust depth interval from the depths of the valid latents projected into the camera coordinate system, denoting by $d_{\mathrm{lo}}$ and $d_{\mathrm{hi}}$ the 1\% and 99\% quantiles of the valid depths. In \cref{eq:decoder_init}, we aggregate the depths of neighboring latents according to the aggregation weights to produce a depth prior $z^{prior}$ for each pixel, and normalize it.

\begin{equation}
  u_v^{\mathrm{prior}}
  =
  \operatorname{clip}
  \left(
  \frac{
  (z_p^{\mathrm{prior}})^{-1}-d_{\mathrm{hi}}^{-1}
  }{
  d_{\mathrm{lo}}^{-1}-d_{\mathrm{hi}}^{-1}
  },
  0,1
  \right).
  \end{equation}

The depth decoder then predicts the correction term  $r_v$ relative to the depth prior.
\begin{equation}
  \hat u_v
  =
  \sigma\left(
  \operatorname{logit}(u_v^{\mathrm{prior}})
  +r_v
  \right),
\end{equation}

and the final depth is
  \begin{equation}
  \hat D_v
  =
  \left[
  d_{\mathrm{hi}}^{-1}
  +
  \hat u_v\cdot
  \left(
  d_{\mathrm{lo}}^{-1}-d_{\mathrm{hi}}^{-1}
  \right)
  \right]^{-1}.
  \end{equation}

\subsection{Camera-Controlled Point Diffusion}
\label{sec:app_point}
  This first stage DiT generates the spatial support of the target scene by predicting the 3D positions of target latent representations. Given
  the reference latent state and a target camera trajectory, the model predicts the target latent positions without directly
  generating their content features.

\textbf{Point DiT}
The Point DiT consists of 6 refinement groups. Each group first performs camera ray cross-attention, followed by two layers of point self-attention and reference memory cross-attention. The point self-attention models the spatial dependencies among the target anchors, while the memory cross-attention transfers geometric and content information from the reference scene. The model has a hidden dimension of 768, uses 8 attention heads, and adopts continuous 3D RoPE.

\textbf{Reference Memory Cross Attention.}
 For each reference latent, we construct a memory token from its 3D position, latent content feature, and geometric descriptor:
\begin{equation}
    \mathbf m_i
    =
    \operatorname{MLP}
    \left(
    \left[
    \phi_{\mathrm{pt}}(\mathbf p_i^{\mathrm{ref}}),
    \mathbf z_i^{\mathrm{ref}},
    \mathbf des_i^{\mathrm{ref}}
    \right]
    \right),
  \end{equation}
where $\mathbf des_i^{\mathrm{ref}}$ contains the viewing ray, log-depth, validity flag, and lattice
  coordinates.

  The noisy features then cross-attend to $m$ to read the reference scene information.

\textbf{Target Camera Cross Attention}
  Target poses are expressed relative to the first view:

   \begin{equation}
    \widetilde{\mathbf T}_v
    =
    \mathbf T_0^{-1}\mathbf T_v,
  \end{equation}
   where $\mathbf T_0$ defines the canonical reference frame. The reference and target point clouds are transformed into this
  coordinate system and standardized using the training-set statistics.

For each future view, we construct a sparse ray map from the VAE feature lattice. Each ray token contains Plücker coordinates,
  image-plane coordinates, trajectory time, and projected depth:

  \begin{equation}
    \mathbf ray_{v,j}
    =
    \left[
    \mathbf l_{v,j},
    \mathbf u_{v,j},
    \boldsymbol{\tau}_{v},
    \ell_{v,j}^{\mathrm{known}},
    m_{v,j}^{\mathrm{known}}
    \right].
  \end{equation}
where $\mathbf l_{v,j}\in\mathbb{R}^{6}$ is the Plücker ray, $\mathbf u_{v,j}\in\mathbb{R}^{2}$ is the normalized lattice
  coordinate, and $\boldsymbol{\tau}_{v}$ encodes the temporal position of the view along the camera trajectory.  
  
  $\ell_{v,j}^{\mathrm{known}}$ denotes the log-depth of the
  closest reference latent projected into the $j$-th ray cell of target
  view $v$. Meanwhile, $m_{v,j}^{\mathrm{known}}$ is a binary validity
  mask specifying whether the cell receives any valid reference projection. Both quantities are derived exclusively from the observed reference point cloud, rather than target-view ground-truth depth. They
  tell the model which parts of a future camera ray are already geometrically constrained and which remain
  unobserved and must be generated.

  For each noisy target point $\hat{p}_i^t$
 , it is first projected onto each future viewpoint to obtain the corresponding query grid coordinates. The point query then interacts with the ray tokens near the projected location via 2D continuous rotary position encoding. The outputs across different future viewpoints are summed rather than combined via cross-view softmax, and are gated by masks indicating the visibility of the point and whether it lies in front of the camera. As a result, the camera trajectory can directly constrain the spatial positions and coverage of the generated point set.

\subsection{Geometric-guided Content Diffusion}
\label{sec:app_content}
  The Content Diffusion takes the 3D layout generated by Point Diffusion as its basis, generates the content features of each latent, and corrects local position errors. This stage no longer receives the target camera trajectory, since camera control has already been converted into explicit 3D geometry in the first stage. Therefore, the second stage mainly addresses geometry-guided content generation.

The model adopts a Wan-style Transformer with 22 layers and a hidden dimension of 1536. Each layer sequentially performs self-attention, reference-state cross-attention, and a feed-forward network. The model employs continuous 3D RoPE in all attention layers: the target self-attention uses the current geometric positions, while the cross-attention uses the target positions and reference positions, respectively, making content interaction explicitly depend on 3D spatial relationships.

For the content features, we add noise according to \cref{eq:content_noise} and use \cref{eq:content_loss} as the supervision loss.

Although Point Diffusion can generate future 3D positions $\hat{P}_{target}$, its generation relies only on reference-frame information and future camera poses, which may lead to erroneous predictions. Since clean content helps correct positions, we construct a correction path from $\hat{P}_{target}$ and $P_{target}$:
\begin{equation}
    \widetilde{p}^t=(1-t_{pos})\cdot P_{target}+t_{pos}\cdot\hat{P}_{target}
\end{equation}
and then predict $P_{target}-\widetilde{p}^t$ based on the content features and positions.

Note that the content time step $t$
  and the geometry time step $t_{pos}$
  are sampled independently. As a result, the model may observe combinations such as "content highly noisy but geometry accurate" or "content nearly clean but geometry still erroneous." This prevents content generation from implicitly judging geometric quality based on content noise, and forces it to read the current 3D positions.

At inference time, the geometry is initialized with the positions from $\hat{P}^{target}$, and the content is initialized from Gaussian noise. The model first performs content denoising, and geometry correction is carried out synchronously during the last 10 content sampling steps. This is to ensure that position corrections are predicted only when the content is relatively accurate. Furthermore, we additionally predict a confidence score to determine whether the geometric position of the current point requires strong correction. In this way, latent positions that are already relatively accurate are preserved as much as possible, while low-confidence latents receive stronger position corrections. The updated positions continue to be used for the 3D RoPE in the next step, allowing the progressively generated content features to in turn guide local geometry optimization.

\section{More Experiments}
\label{sec:app_exp}

\begin{figure}[t]
    \centering
    \includegraphics[width=1.0\linewidth]{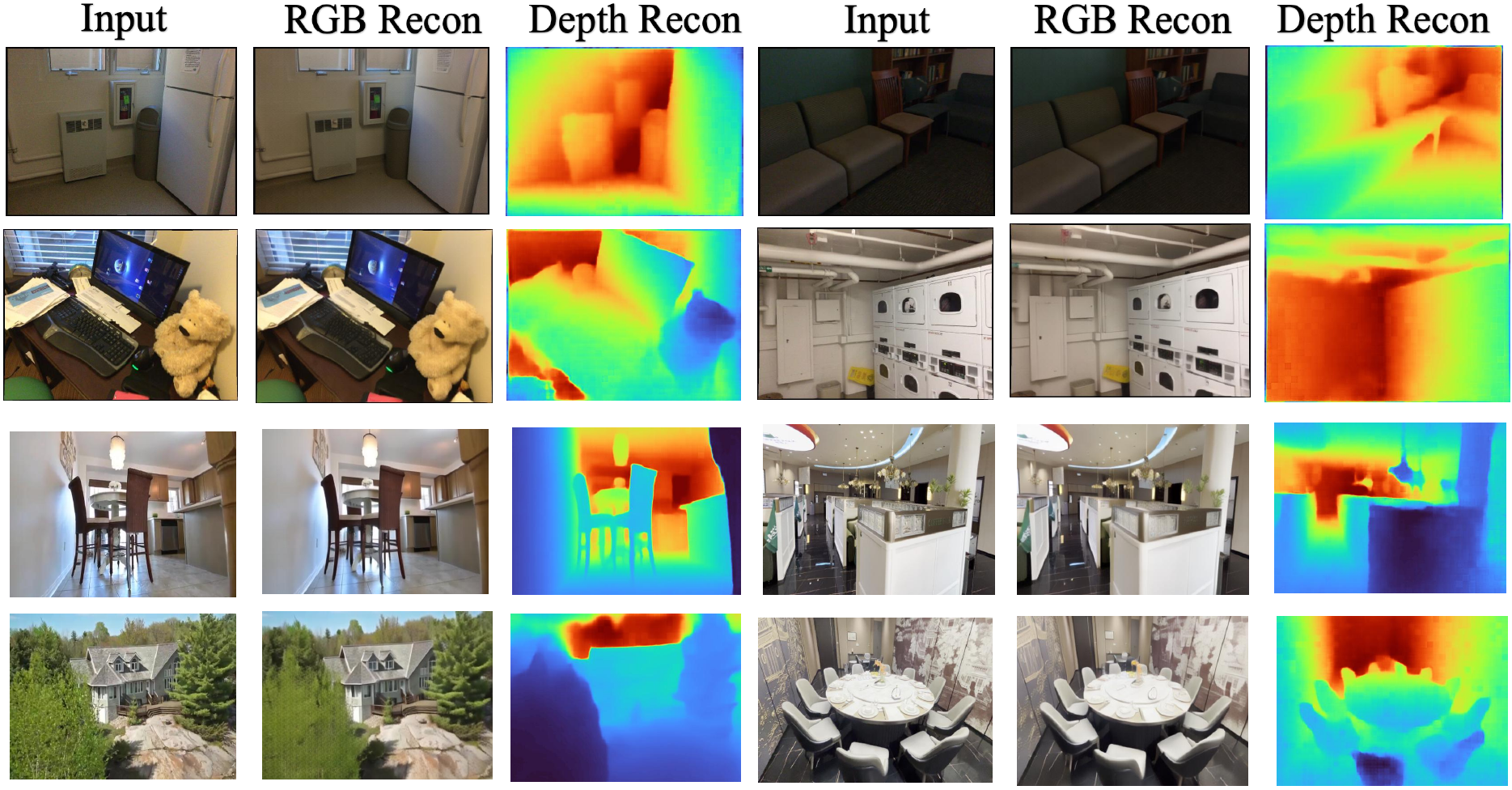}
    \vspace{-3mm}
    \caption{Visualization of the Decoder Reconstruction.}
    \label{fig:sp_fig1}
    \vspace{-5mm}
\end{figure}

\subsection{Latent Decoder Reconstruction}
\label{sec:app_recon} 
In this part, we test the Decoder reconstruction results for the 3D latents encoded by the Encoder.

\textbf{Visualization of RGB and Depth Reconstruction.}
\Cref{fig:sp_fig1} visualizes the RGB and Depth results reconstructed by the Decoder. Although we use only sparse and unstructured 3D latents, through our designed Decoder we can still reconstruct high-fidelity RGB and Depth observations.

\begin{table}[t]
  \centering
  \begin{minipage}[b]{0.485\textwidth}
    \centering
    \caption{Ablation of dual-path deformable cross-attention on DL3DV.}
    \label{sp_tab3}
  \end{minipage}\hfill
  \begin{minipage}[b]{0.485\textwidth}
    \centering
    \caption{Ablation of vision encoders on DL3DV.}
    \label{sp_tab4}
  \end{minipage}
  \par\vspace{-1mm}
  \footnotesize
  \begin{minipage}[t]{0.485\textwidth}
    \centering
    \setlength{\tabcolsep}{3pt}
    \begin{tabularx}{\linewidth}{c|YYYY}
      \toprule
      \multirow{2}{*}{Path} & \multicolumn{2}{c}{RGB} & \multicolumn{2}{c}{Depth} \\
      & PSNR$\uparrow$ & SSIM$\uparrow$ & Acc.$\downarrow$ & Comp.$\downarrow$ \\
      \midrule
      Main-only & 14.9 & 0.43 & 0.49 & 0.70 \\
      MV-only & 14.0 & 0.36 & 0.38 & 0.59 \\
      \midrule
      \rowcolor{blue!10}
      Main+MV & \textbf{15.4} & \textbf{0.45} & \textbf{0.34} & \textbf{0.55} \\
      \bottomrule
    \end{tabularx}
  \end{minipage}\hfill
  \begin{minipage}[t]{0.485\textwidth}
    \centering
    \setlength{\tabcolsep}{3pt}
    \begin{tabularx}{\linewidth}{c|YYYY}
      \toprule
      \multirow{2}{*}{Encoder} & \multicolumn{2}{c}{RGB} & \multicolumn{2}{c}{Depth} \\
      & PSNR$\uparrow$ & SSIM$\uparrow$ & Acc.$\downarrow$ & Comp.$\downarrow$ \\
      \midrule
      DINO & 14.5 & 0.39 & 0.43 & 0.66 \\
      VAE & 14.4 & 0.37 & 0.52 & 0.71 \\
      VGGT & 15.1 & 0.42 & 0.39 & 0.58 \\
      \midrule
      \rowcolor{blue!10}
      DA3-Base & \textbf{15.4} & \textbf{0.45} & \textbf{0.34} & \textbf{0.55} \\
      \bottomrule
    \end{tabularx}
  \end{minipage}
  \vspace{-4mm}
\end{table}

\subsection{Ablation Study of Latent Encoder-Decoder.}
\textbf{Dual Path Deformable Cross Attention.} 
In this part, we test the design of Dual Path Deformable Cross Attention in the latent encoder. We compare the performance of using only the Main Path, only the Multi-View Path, and the combination of both, with the results shown in \Cref{sp_tab3}. As we analyzed, the Main Path provides accurate local texture information by focusing only on regions from depth-consistent frames, and thus achieves better RGB-related metrics. However, due to the lack of information across multiple views, its geometric features are weaker, resulting in worse Depth-related metrics. The Multi-View Path, through aggregation across multiple views, effectively provides accurate geometric features, but may be affected by occlusion and other factors that introduce erroneous texture information, leading to better Depth metrics but weaker RGB metrics. By combining the two, we can simultaneously achieve superior RGB and Depth generation.

\textbf{Different Vision Encoders.} Here we test the performance comparison of different visual foundation models when used as the frozen encoder. Specifically, we compare DINOV2~\citep{dinov2}, VAE~\citep{vae}, VGGT, and the DA3-Base adopted by the existing implementation. It should be noted that since DINOV2 and VAE cannot directly output depth, a learnable depth prediction head must be attached. The results are shown in \Cref{sp_tab4}. We can see that VAE performs the worst because it focuses only on texture information and lacks an understanding of geometry, resulting in poor performance in both depth prediction and subsequent depth reconstruction. DA3-Base matches VGGT at lower computational cost, so we adopt it.

\begin{figure}[t]
    \centering
    \includegraphics[width=1.0\linewidth]{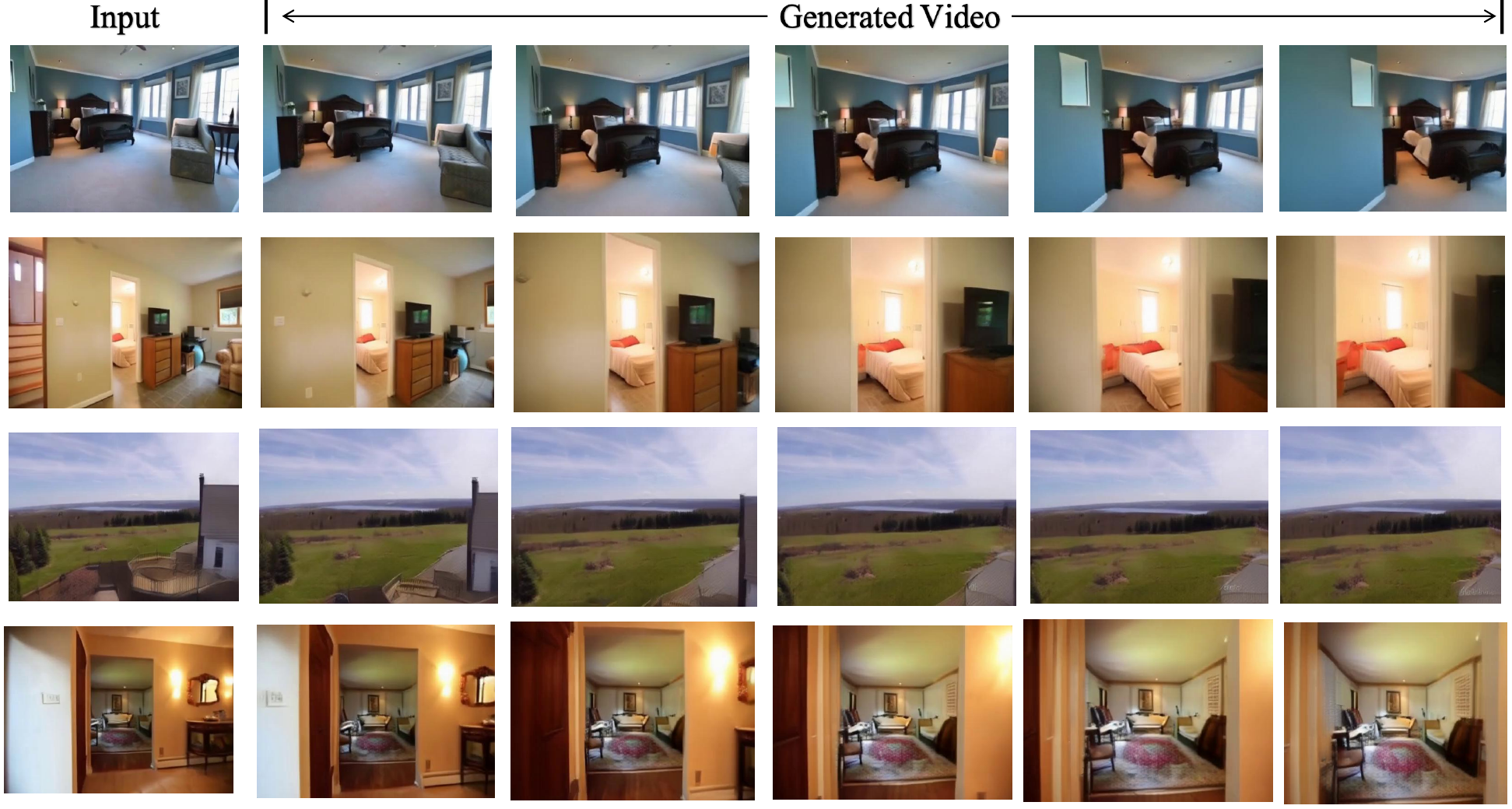}
    \vspace{-3mm}
    \caption{Visualization in the 1-view setting with camera condition.}
    \label{fig:sp_3}
    \vspace{-5mm}
\end{figure}

\begin{figure}[t]
    \centering
    \includegraphics[width=1.0\linewidth]
    {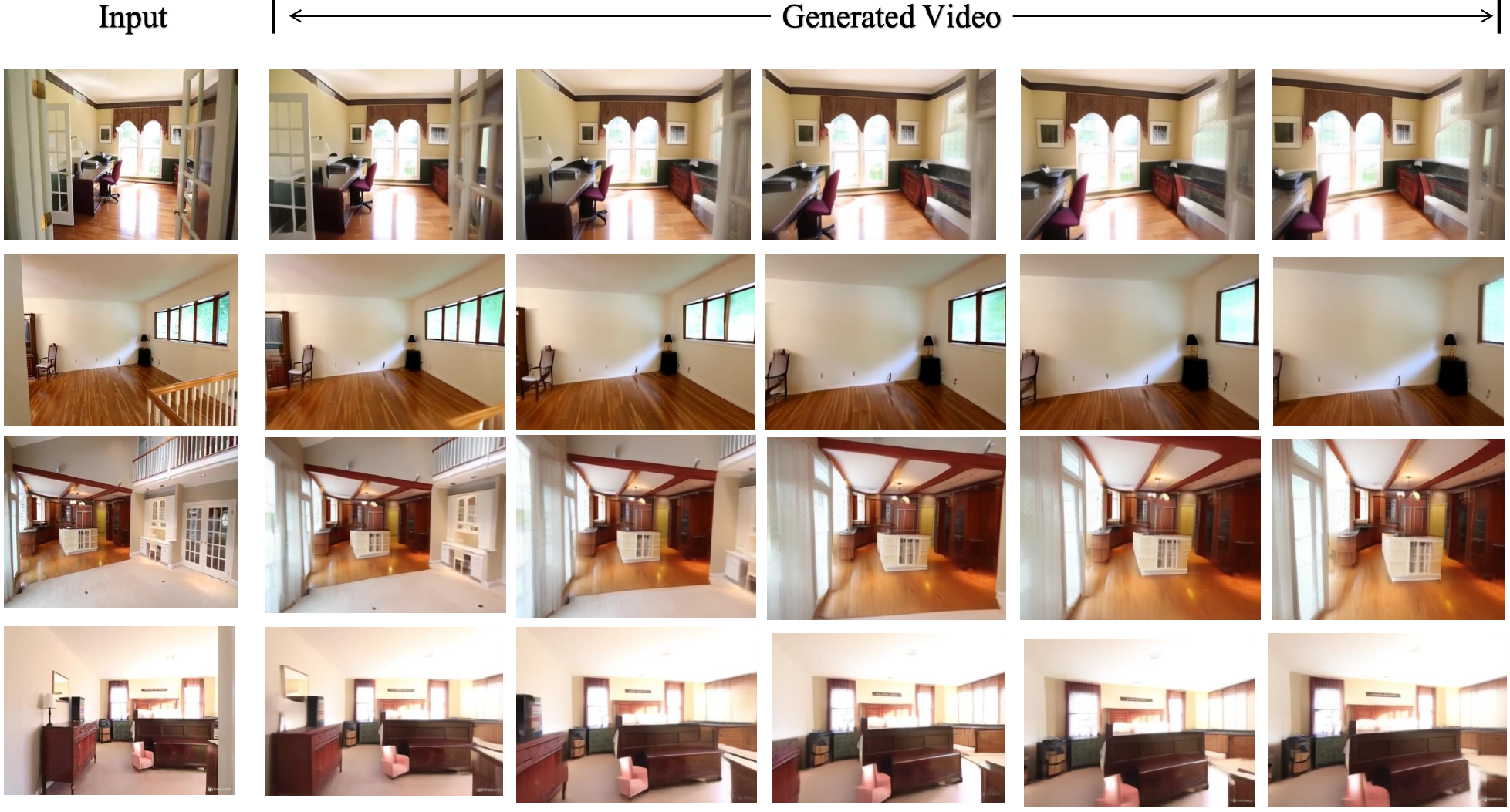}
    \vspace{-3mm}
    \caption{Visualization in the 1-view setting without camera condition.}
    \label{fig:sp_4}
    \vspace{-5mm}
\end{figure}
\subsection{3D Scene Generation}
\label{sec:app_gen}
\textbf{Visualization of 1-View Setting.} We visualize generation results under the 1-view setting. \Cref{fig:sp_3} shows results with camera control, while \Cref{fig:sp_4} shows results without camera control.

\begin{table}[t]
    \centering
    \caption{Quantitative Comparison of Appearance Generation
on Out-of-Distribution Datasets with 1-view setting. We report 2D metrics (PSNR, SSIM) and 3D metrics (Accuracy, Completeness).}
    \vspace{-2mm}
    \label{sp_tab5}
    \resizebox{1.0\linewidth}{!}{
    \begin{tabular}{c|cccccccc}
    \toprule
    \multirow{2}{*}{Method}&\multicolumn{4}{c}{TartanAir}&\multicolumn{4}{c}{ScanNet++}\\
         &PSNR&SSIM&Accuracy&Completeness&PSNR&SSIM&Accuracy&Completeness\\
         \midrule
         Aether&12.1&0.43&0.55&0.81&13.4&0.45&0.50&0.71\\
         Gen3R&13.9&0.47&0.52&0.74&14.2&0.49&0.46&0.70\\
         \midrule
         \rowcolor{blue!10}
         Ours&15.2&0.53&0.22&0.49&15.8&0.55&0.16&0.37\\
         \midrule
    \end{tabular}}
    \vspace{-4mm}
\end{table}

\textbf{Out-of-Distribution Datasets.} 
In this section, we test the generation performance on datasets outside the training distribution (TartanAir and ScanNet++) to demonstrate the generalization of our method, with the results shown in \Cref{sp_tab5}. We can see that even on out-of-distribution datasets, our method still achieves superior performance, indicating strong generalization.

\begin{table}[t]
  \centering
  \begin{minipage}[b]{0.485\textwidth}
    \centering
    \caption{Ablation of the noisy input of Point Diffusion.}
    \label{sp_tab6}
  \end{minipage}\hfill
  \begin{minipage}[b]{0.485\textwidth}
    \centering
    \caption{Ablation of position correction in Content Diffusion.}
    \label{sp_tab7}
  \end{minipage}
  \par\vspace{-1mm}
  \footnotesize
  \begin{minipage}[t]{0.485\textwidth}
    \centering
    \setlength{\tabcolsep}{3pt}
    \begin{tabularx}{\linewidth}{l|YYYY}
      \toprule
      \multirow{2}{*}{Noise} & \multicolumn{2}{c}{DL3DV} & \multicolumn{2}{c}{RE10K} \\
      & PSNR$\uparrow$ & SSIM$\uparrow$ & PSNR$\uparrow$ & SSIM$\uparrow$ \\
      \midrule
      Gaussian & 14.4 & 0.37 & 15.9 & 0.48 \\
      \midrule
      \rowcolor{blue!10}
      Reference & \textbf{15.4} & \textbf{0.45} & \textbf{17.1} & \textbf{0.60} \\
      \bottomrule
    \end{tabularx}
  \end{minipage}\hfill
  \begin{minipage}[t]{0.485\textwidth}
    \centering
    \setlength{\tabcolsep}{3pt}
    \begin{tabularx}{\linewidth}{l|YYYY}
      \toprule
      \multirow{2}{*}{Correction} & \multicolumn{2}{c}{DL3DV} & \multicolumn{2}{c}{RE10K} \\
      & PSNR$\uparrow$ & SSIM$\uparrow$ & PSNR$\uparrow$ & SSIM$\uparrow$ \\
      \midrule
      None & 14.9 & 0.40 & 16.4 & 0.53 \\
      w/o confidence & 15.1 & 0.41 & 16.7 & 0.58 \\
      \midrule
      \rowcolor{blue!10}
      w/ confidence & \textbf{15.4} & \textbf{0.45} & \textbf{17.1} & \textbf{0.60} \\
      \bottomrule
    \end{tabularx}
  \end{minipage}
  \vspace{-4mm}
\end{table}

\textbf{Ablation Studies.} In this part, we test the noise-adding schemes of Point Diffusion. Specifically, Gaussian Noise denotes the standard noise-adding scheme, i.e., constructing a flow path from Gaussian noise to the target point cloud. However, considering the sensitivity of point cloud positions, we adopt a new noise-adding scheme, namely a flow path from the reference-frame point cloud to the target point cloud (Reference Noise), which maximally stabilizes point cloud generation and effectively leverages the initial information of the scene. The results are shown in \Cref{sp_tab6}. We can see that Reference Noise achieves higher performance, validating our design.

We then test geometry correction in content diffusion (\Cref{sp_tab7}). It can be seen that introducing geometry correction effectively improves performance while bringing almost no additional overhead, indicating that texture content also plays a supporting role for geometric structure.

\begin{figure}[t]
    \centering
    \includegraphics[width=1.0\linewidth]{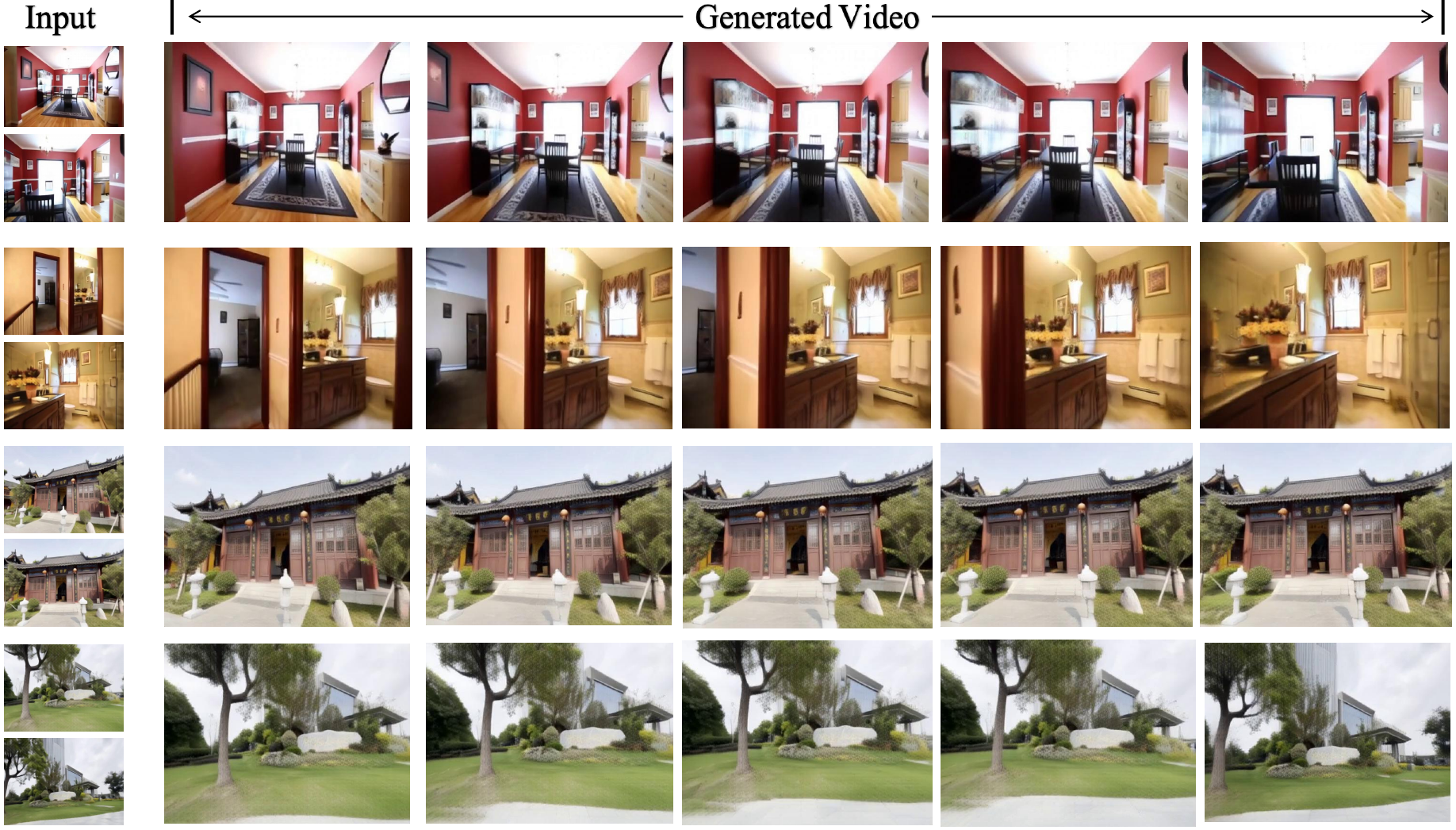}
    \vspace{-3mm}
    \caption{Visualization in the 2-view setting with camera condition.}
    \label{fig:sp_5}
    \vspace{-5mm}
\end{figure}

\begin{figure}[t]
    \centering
    \includegraphics[width=1.0\linewidth]{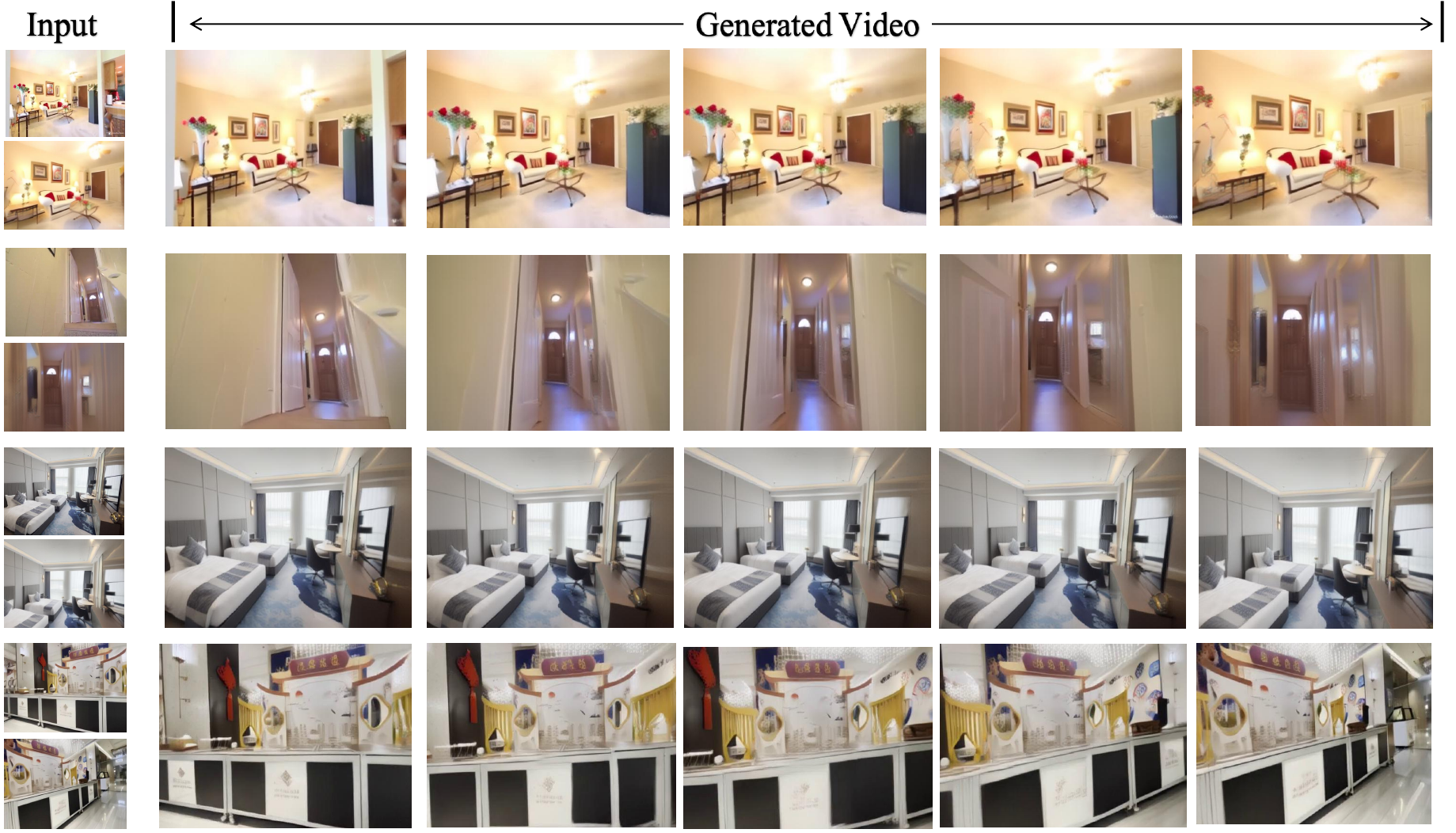}
    \vspace{-3mm}
    \caption{Visualization in the 2-view setting without camera condition.}
    \label{fig:sp_6}
    \vspace{-5mm}
\end{figure}

\textbf{Visualization of 2-View Setting.} We visualize generation results under the 2-view setting. \Cref{fig:sp_5} shows results with camera control, while \Cref{fig:sp_6} shows results without camera control.

\section{Discussion}
\subsection{Difference with Feed-Foward 3D Gaussian Splatting}
Feed-forward 3D Gaussian methods and our approach both construct explicit 3D scene representations without per-scene optimization,
  but they differ substantially in how scene information is organized and used. Many feed-forward 3DGS methods predict one or
  multiple Gaussian primitives from each input pixel or image patch. Although these primitives are subsequently transformed into a
  shared 3D coordinate system, their prediction remains aligned with the source observations. Consequently, the same physical
  surface observed from multiple viewpoints may be represented by several overlapping Gaussians. This redundancy is beneficial for
  preserving view-specific details, but it also causes the representation size to grow with the number and resolution of input
  views.

Such pixel-aligned representations are well suited to reconstruction. Each Gaussian explicitly stores the attributes required for
  rendering, including position, spatial extent, opacity, and appearance. Once the Gaussians have been predicted, novel views can be
  rendered efficiently through Gaussian rasterization. The dense correspondence between image pixels and Gaussian primitives also
  helps retain fine textures and thin structures, allowing fast and high-fidelity
  reconstruction.

  However, the same design is less convenient as a latent space for generative scene modeling. First, redundant primitives require a
  generative model to repeatedly model the same physical region across different view-aligned Gaussian predictions. Second, the
  number of primitives is large, increasing the
  computational cost of diffusion or autoregressive generation. Third, each primitive jointly carries geometry, scale, opacity, and
  appearance attributes. Generating this high-dimensional and coupled parameter space requires the model to
  resolve scene layout, visibility, surface extent, and texture.

In contrast, we construct a compact, scene-level 3D representation specifically for generative modeling. By representing shared
  scene regions only once, our latent space reduces both the representation size and the burden of generative learning. Moreover,
  the separation between explicit geometry and latent content allows camera-controllable generation to be decomposed into geometry
  generation followed by content generation, providing a more direct control path. The trade-off is that our representation relies
  on a learned decoder and therefore does not match the rendering speed or fine-detail fidelity of rasterized 3D Gaussians.
  Overall, feed-forward 3DGS favors efficient, high-fidelity reconstruction, whereas our representation favors compact and
  controllable scene generation.

  \subsection{Limitations}
\textbf{Rendering Fidelity.} Compared with conventional video VAEs, our representation provides lower reconstruction sharpness. Dense
  video-VAE latents preserve spatially aligned appearance information at every view, whereas our method compresses the scene into a
  limited number of sparse 3D latents and reconstructs each view through a learned decoder. This stronger information bottleneck may
  smooth high-frequency textures and lose details in thin structures, object boundaries, and view-dependent appearance. Improving
  the spatial adaptivity of the anchors and the capacity of the rendering decoder is a promising direction for closing this fidelity
  gap.

\textbf{Dynamic-scene Modeling.} Our current representation assumes that scene content can be anchored to persistent locations in a shared
  3D coordinate system. This assumption is effective for static scenes but becomes restrictive for moving or
  deformable objects. Observations of a dynamic object at different times may correspond to different spatial locations, making
  direct multi-view aggregation ambiguous and potentially leading to duplicated geometry or inconsistent appearance. Moreover, the
  generative model predicts spatial scene completion without explicitly modeling object motion or temporal evolution.
  Extending the representation with time-dependent anchors, motion fields, or explicit static-dynamic decomposition is
  necessary for general dynamic environments.

\end{document}